%% file: manuscript.tex
\documentclass[review]{elsarticle}
\usepackage{lineno,hyperref}
\modulolinenumbers[1]
\usepackage{booktabs}
\usepackage{array}
\usepackage[utf8]{inputenc}

\usepackage{graphicx}

\usepackage{adjustbox}
\usepackage{subcaption}
\usepackage{float}
\usepackage{placeins}
\usepackage{stfloats}

\usepackage{xcolor}

\usepackage{amsfonts,amsmath}

\usepackage{amsthm}
\usepackage{enumerate} 
\usepackage{graphicx} 

\usepackage{epstopdf} 

\usepackage{todonotes}
\usepackage{comment}
\usepackage{mathtools}
\usepackage{cancel}
\usepackage{dsfont}
\def\submission{0}
\if\submission1
\usepackage[doublespacing]{setspace}
\usepackage{lineno}
\linenumbers
\fi

\newcommand{\R}{\mathbb{R}}

\newcommand{\Ind}{\mathds{1}}

\newcommand{\centerpix}{c}
\newcommand{\window}{W}
\newcommand{\voxel}{v}
\newcommand{\pixel}{v}
\newcommand{\image}{I}

\usepackage{lipsum}
\makeatletter
\def\ps@pprintTitle{%
 \let\@oddhead\@empty
 \let\@evenhead\@empty
 \def\@oddfoot{}%
 \let\@evenfoot\@oddfoot}
\makeatother
\begin{document}
\begin{frontmatter}

\title{Determination of droplet size from wide-angle light scattering  image data using convolutional neural networks}

\author[]{Tom Kirstein$^1$,
    Simon Aßmann$^2$,  Orkun Furat$^1$,
    Stefan Will$^{2}$
    and Volker Schmidt$^1$}

\address{
$^1$Institute of Stochastics, Ulm University,
   89069  Ulm, Germany}
    \address{
$^2$Institute of Engineering Thermodynamics (LTT) and Erlangen Graduate School in Advanced Optical Technologies (SAOT), Friedrich-Alexander-Universität Erlangen-Nürnberg, 91058 Erlangen, Germany\\
  Corresponding Author: Tom Kirstein, tom.kirstein@uni-ulm.de}

\begin{abstract}
Wide-angle light scattering (WALS) offers the possibility of a highly temporally and spatially resolved measurement of droplets in spray-based methods for nanoparticle synthesis. The size of these droplets is a critical variable affecting the final properties of synthesized materials such as hetero-aggregates. However, conventional methods for determining droplet sizes from WALS image data are labor-intensive  and may introduce biases, particularly when applied to complex systems like spray flame synthesis (SFS). To address these challenges, we introduce a fully automatic machine learning-based approach that employs convolutional neural networks (CNNs) in order to streamline the droplet sizing process. This CNN-based methodology offers further advantages: it requires few manual labels and can utilize transfer learning, making it a promising alternative to conventional methods, specifically with respect to efficiency. 
To evaluate the performance of our machine learning models, we consider WALS data from an ethanol spray flame process at various heights above the burner surface (HABs), where the models are trained and cross-validated on a large dataset comprising nearly 35000 WALS images.


\end{abstract}

\end{frontmatter}

\section{Introduction}

    In modern nanotechnology, the significance of spray-based methods for tailored nanoparticle synthesis is paramount, serving as a key for advancing materials science. Notably, with techniques such as spray flame synthesis (SFS) \cite{Assmann2024} and oppositely charged electrosprays (ES) in combination with high-temperature furnace systems \cite{Tang2017}, not only the production of diverse ceramic nanoparticle systems such as titania, silica, alumina and iron oxides is feasible, but also of so-called hetero-aggregates. Such particle systems, composed of distinct materials with defined interface-contacts, may introduce novel electronic, mechanical, and optical properties that transcend those of their individual constituents. For example, a noteworthy application in the realm of photocatalysis involves hetero-aggregates formed by titanium dioxide (\( \text{TiO}_2 \)) and tungsten trioxide. The interface-contacts in these hetero-aggregates play a crucial role in spatially separating photogenerated electron-hole pairs. This separation minimizes their direct recombination, thereby enhancing the photocatalytic performance beyond that of pure \( \text{TiO}_2 \) \cite{Low2017,PinedoEscobar2021,Kwon2000}.

    For the production of hetero-aggregates by SFS, the control of the atomization and evaporation processes is  crucial as the resulting droplet size distributions profoundly influence the final nanoparticle characteristics, such as size, morphology, and composition. Therefore, a sophisticated \textit{in situ} measurement of droplet sizes and their distribution is required for a detailed understanding of the underlying processes and a controlled particle synthesis \cite{Maedler}.

    The investigation of droplet sizes hereby requires noninvasive laser-based measurement techniques as they provide high temporal and/or spatial resolution. Phase Doppler anemometry (PDA) is commonly used in fluid dynamics research and industrial applications to determine statistical distributions of the size and velocity of droplets. Fast photodetectors allow the rapid detection of thousands of individual droplets in comparably short measurement times of a few seconds to minutes~\cite{Lichti_2018}. However, accurate droplet sizing is limited to homogeneous and spherical objects and thus prone to errors when investigating processes such as the SFS, where evaporating droplets and already synthesized nanoparticles of various shapes might be present in the measurement volume simultaneously. Here, the wide-angle light scattering (WALS) approach~\cite{Huber_2016,Oltmann_2010} is favorable as it allows the detection of almost continuous scattering patterns and hence a separation of the different angular scattering patterns from nanoparticles and micrometer-sized droplets~\cite{Assmann2024,Assmann2020,Assmann2021}.  
    
    Aßmann et al.~\cite{Assmann2021} developed an evaluation algorithm for droplet sizing based on the characteristic local maxima in the Mie scattering patterns from homogeneous spheres with diameters above  1 $\mu$m. In this way, the determination of droplet size distributions is possible even with a smooth scattering background from nanoparticles allowing a comprehensive in situ investigation of SFS processes with respect to both droplet evolution and nanoparticle formation \cite{Assmann2021}. However, the detection of the local maxima via the parameterized MATLAB-based function in WALS data might lead to biased or erroneous results when investigating different SFS systems or sprays. Moreover, a manual and time-consuming verification of a correct detection of local maxima is required for validation. Here, the implementation of convolutional neural networks (CNN) might improve the evaluation of droplet sizes from WALS data. Specifically, by employing CNNs, we can provide a more efficient method for the prediction of droplet sizes in WALS images that requires less manual labeling. Such CNN-based approaches have been applied to a variety of different tasks, e.g,  the efficient labeling of image data, see~\cite{Unet}, or  image classification and regression tasks, see~\cite{Laue,Samy,Yadav}.   While first attempts to apply  machine learning algorithms for the evaluation of scattering data from nanoparticles and heterogenous materials were successful~\cite{saxs,Talebi_2020}, a similar approach for the evaluation of scattering from single droplets has not been applied so far.

    Therefore, in this study, we  investigate the potential of machine learning techniques, specifically CNNs, to predict the size of droplets in WALS images. 
    To test the effectiveness of this approach, we utilize droplets observed in WALS images that are the result of a spray flame process with ethanol at heights above burner (HABs) ranging from 20 mm to 120 mm. This range provides a useful testbed for evaluating the performance of our machine learning-based approach. 
    
    The rest of this paper is organized as follows. In Section~\ref{sec.two} 
    we describe the WALS data as well as the CNN-based methods, which we use to predict the droplet sizes.
    The results of our analysis, including an evaluation of how well the predictions allow to derive a droplet size distribution, are presented in Section~\ref{sec.the}.
    In Section~\ref{sec.fou} the obtained results are discussed. Finally, Section~\ref{sec.fiv} concludes and gives an outlook to possible future research.

\section{Materials and methods}\label{sec.two}

 In this section, we present our data and methods in detail. In particular, we describe how we applied CNNs to predict the size of droplets in WALS images.

\subsection{Image data acquisition}\label{sec:imdat}

\begin{figure}[ht!]
        \centering
        \includegraphics[width=7cm]{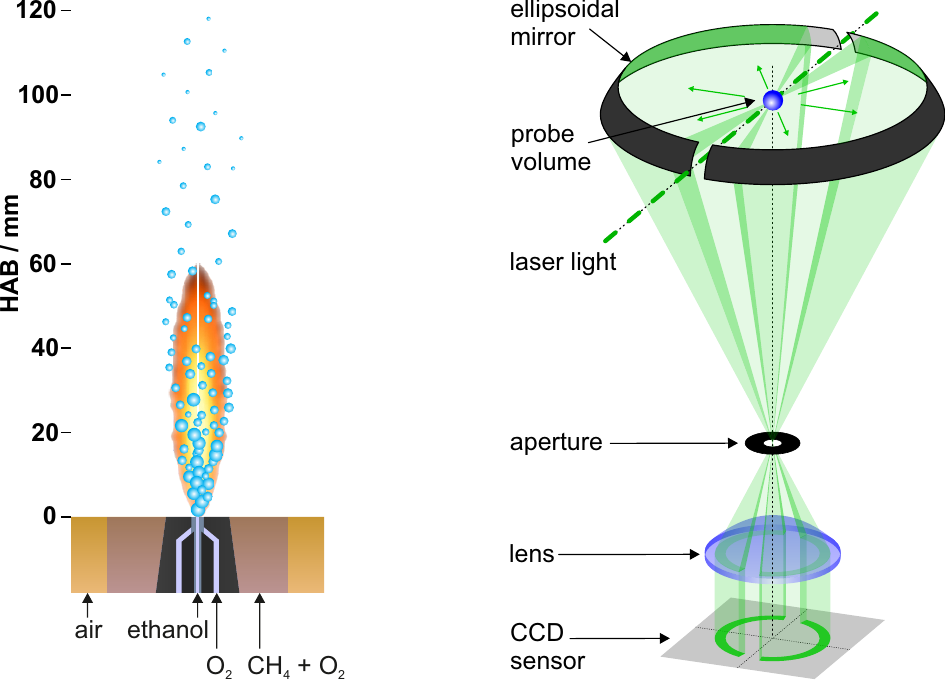}
        \caption{Spray flame burner (\textit{SpraySyn}) (left) and WALS measurement principle (right)}
        \label{fig:SFS_WALS}
    \end{figure}

    The experimental setup for the spray flame burner (\textit{SpraySyn}), its operation and the WALS measurements system is described in detail in \cite{Assmann2020, Assmann2021,Schneider_2019}, Thus, here only a brief summary of the spray flame burner and the acquisition of scattering data utilizing the WALS approach is given. Schemes of both the SpraySyn burner and the WALS measurement principle, respectively, are depicted in Figure~\ref{fig:SFS_WALS}.

    The SpraySyn burner features a two-fluid nozzle, which atomizes a (precursor-laden) solution with a dispersion gas into a hot pilot flame. The nozzle is surrounded by a bronze sinter-matrix that directs the premixed combustion gas for the pilot flame inward and shields the flame using an inert sheath gas. Flow rates for gases are controlled by mass flow controllers and set to 10 slm oxygen for atomization, a mixture of 2 slm methane and 16 slm oxygen for the pilot flame and 120 slm pressurized and dried air for the sheath gas. Pure ethanol (EtOH) is used as solvent and pumped through the inner capillary at 2 ml/min by a syringe pump.
    
    The essential component of WALS measurements is an ellipsoidal mirror with two focal points at a defined distance. In the first focal point, the probe volume, objects such as droplets are irradiated by a beam of monochromatic ($\lambda$ = 532 nm) and vertically polarized laser light guided through two slits on each half of the mirror. The elastically scattered light is captured by the mirror surface in the horizontal scattering plane and directed through an aperture in the second focal point of the mirror and a camera lens (\textit{f} = 12.5 mm) onto a CCD sensor. The diameter of the laser beam (1.0 mm) and the aperture of the camera (\textit{f}/2.0) define the size of the cylindrically shaped probe volume (length~$\approx$~12 mm) and thus the spatial region from which scattered light is recorded by the camera. The maximum angular resolution ($\approx$~0.2°) is determined by the pixel size of the CCD sensor of the equipped camera (Pike F-100 B, Allied Vision Technologies GmbH, $1000\times 1000$ pixel) and the maximum angular region (10° to 170°) by the two slits in the mirror through which the laser beam is guided, resulting in a WALS measurement image for which the scattering information is located in two ring-shaped segments, see, for example, Figure~\ref{fig:preprocessing} (non-black region). 
    
    In this paper, we reused the WALS measurement data acquired from the EtOH spray flame at heights above burner surface (HAB) between 20 mm and 120 mm (10 mm step) \cite{Assmann2020}. At each HAB between 2000 and 6000 scattering images (nearly 35000 in total) were acquired and serve as a database for the training and evaluation of the  CNN-based approach described in Section~\ref{sec.cnn.app}.

\subsection{Generation of training data}\label{sec:gt}

To effectively train neural networks for the purpose of predicting droplet sizes from WALS image data, it is essential to obtain accurate ground truth labels, i.e., the size of a droplet observed in a WALS image. These labels were determined using the algorithm introduced in~\cite{Assmann2021} for each WALS image considered in the present paper.  Importantly, each WALS image is divided into two ring-shaped segments, left and right, and droplet sizes are initially determined independently for each. 

The division of each WALS image into two separate ring-shaped segments, left and right, allows for more precise measurements in some scenarios ,e.g., when multiple droplets enter the probe volume or a measured droplet is significantly off-center. In these cases, the signals on the right- and left-hand sides may differ, leading to inaccuracies in size predictions if the entire ring were considered as a whole. By independently determining droplet sizes for each segment, these potential discrepancies can be identified more efficiently.  

More specifically, the first step of this process involves converting the 2D WALS image data   into scattering data using the method described in~\cite{Huber_2016}. In this context, scattering data refers to a 1D representation of each ring-shaped segment of the original 2D WALS image, where each point in the 1D data corresponds to an angle with respect to the center point in the 2D image and has an intensity value that is the average intensity over the corresponding angle in the ring-shaped segment. 

A peakfinder algorithm is then applied to this scattering data in order to isolate peaks that exceed a specific intensity threshold. By analyzing simulated scattering data, a correlation was established between the distance separating these peaks and the size of the droplets, i.e., allowing us to predict the droplet size from the distance between isolated peaks in the scattering data.  After droplet sizes were determined using this correlation, images leading to erroneous detection of local maxima are manually identified, see~\cite{Assmann2020,Assmann2021} for details.

Even though  the droplet size for such erroneously detected images is unknown, they can still be utilized during the training of the neural network. More specifically, two separate networks are trained. The first network is a binary classifier that checks if a droplet can be detected using the  algorithm mentioned above, while the second network estimates the size of the detected droplet. The binary classifier network, which  decides if a droplet  size can be detected, is  trained using images for which droplet size is available and those for which it is not. 

 Using the algorithm described above, each experimentally measured WALS image is assigned two scalar values $s^*_\mathrm{L},s^*_\mathrm{R}$, corresponding to the  droplet sizes determined for the left and right ring-shaped segments, respectively. In cases where no droplet size can be determined, the placeholder value of $-1$ is assigned instead. In order to simplify the subsequent analysis, the resulting vector $(s^*_\mathrm{L},s^*_\mathrm{R})$ is converted into a ground truth vector $(s_1,s_2,s_3,s_4)$, where $s_1=\Ind(s^*_\mathrm{L}>0)$, $s_2=\Ind(s^*_\mathrm{R}>0)$ and $(s_3,s_4)=(s^*_\mathrm{L},s^*_\mathrm{R})$. Here,  
$\Ind$ denotes the indicator of the given condition, which outputs 1 if the condition is true and 0 otherwise.

\subsection{CNN-based approach for the estimation of droplet size}
\label{sec.cnn.app}

In this section, we detail a machine learning-based approach for the automatic prediction of droplet sizes from WALS image data using a CNN. Our goal is to predict the entire ground truth vector $s=(s_1,s_2,s_3,s_4)$ introduced above, which includes both the droplet sizes $s_3,s_4$ and whether or not a droplet was detected, as indicated by the categorical variables $s_1,s_2$. Eventually, our goal is to consolidate this vector $s$, or its estimations, to determine a single, accurate droplet size for each WALS image. Details on this can be found in  Equation~\ref{eq:size} and Section~\ref{sec.the}.

Compared to the procedure used to generate the ground truth, see Section~\ref{sec:gt}, the CNN-based approach offers several advantages. Most importantly, it is fully automatic and does not require manual intervention, which can be time-consuming and prone to errors. Therefore, given the necessity to analyze vast amounts of data, the CNN-based approach can help to streamline the synthesis of complex particles, such as hetero-aggregates. Additionally, if a suitable ground truth is given consisting of additional descriptors, other than the droplet size, the approach can easily be extended in order to predict such additional descriptors.

However, before using the WALS image data for training a  CNN, it is essential to preprocess the data. This preprocessing step is crucial for enabling the CNN to efficiently process the data and accurately predict the ground truth vector. In the following, we will outline in detail how this preprocessing step is performed and how it helps to improve the performance of the CNN-based approach.

\input{false_color}

\subsubsection{Data preprocessing}

The WALS image data can be  described as a map  $I:\window\rightarrow\R,$ where the pixel space $\window  =\{0,1,\dots,999\}\times \{0,1,\dots,999\}$ is a discretized square and
$\image(\pixel)\in\R$ denotes the grayscale value of  pixel $\pixel=(\pixel_1,\pixel_2)\in\window$. 
However, due to the nature of the WALS device, only those pixels within a ring around a pole $(\centerpix_1,\centerpix_2)\in [0,999]\times[0,999]$ are illuminated,  pixels outside of that ring contain no information with respect to the droplet,  see Figure~\ref{fig:preprocessing} (left). Therefore, several steps were taken to prepare the images for analysis. The most important step involves the application of a polar transformation to the images, which is useful for  unwrapping circular objects (such as WALS image data).

More specifically, the polar transformation is applied to the input image $I$ in order to obtain a transformed image $I_\mathrm{p}:\window_\mathrm{p}\rightarrow\R,$ where the polar pixel space $\window_\mathrm{p}$ is a set with equidistant points that spans the entire two-dimensional polar coordinate system $ (0,500] \times [0,2\pi)$. For that purpose, Cartesian coordinates $(v_1,v_2)\in W$ are transformed into polar coordinates $(r,\theta) \in (0,500] \times [0, 2\pi)$ with respect to a given center point $(c_1,c_2)\in[0,999]\times[0,999]$. More precisely, the polar coordinates $(r,\theta)$ are calculated from the Cartesian coordinates $(v_1,v_2)$  using the following equations:
\vspace{-0.5em}
\begin{align*}
r(\voxel_1,\voxel_2) &= \sqrt{(\voxel_1-\centerpix_1)^2 + (\voxel_2-\centerpix_2)^2} \\
\theta(\voxel_1,\voxel_2) &= \tan^{-1}\left(\frac{\voxel_2-\centerpix_2}{\voxel_1-\centerpix_1}\right).
\end{align*}
In this manner we obtain the pair of polar coordinates $(r(\voxel_1,\voxel_2),\theta(\voxel_1,\voxel_2))$ and the corresponding intensity value $I(\voxel_1,\voxel_2)$ for each pixel $(\voxel_1,\voxel_2) \in \window$. The intensity values of the transformed image $I_\mathrm{p}$, which are defined on the set $\window_\mathrm{p}$, are then determined by linear interpolation.

In order to maintain a high image quality while also reducing the image size, the size of the polar transformed image is chosen to be $500\times500.$ Furthermore, in order to determine the region of interest in the polar transformation of $I$  (i.e., the ring in which WALS signals are measured), a reference image was additionally transformed. 
More precisely, this reference image was obtained by using a separate image in which the entire mirror area was well-lit, highlighting the pixels that could potentially contain relevant information. We obtained the range of $r $ and $ \theta $ values that corresponded to the region of interest by identifying the illuminated region in the transformed reference image. Then, the polar transformation $ I_\mathrm{p} $ is cropped to only include this region, thus resulting in a reduced image size of $56\times 496$,  while retaining all relevant information. Note that, in addition to identifying the region of interest, the reference image was also used to obtain the center point exploited for the transformation into polar coordinates.

However, before applying the polar transformation and cropping the images,  a common type of imaging artifacts is identified and removed. These artifacts manifest as singular bright spots with intensity value equal to the maximum intensity value of 255. By “singular,” we mean that these bright spots are isolated and not part of a larger pattern or structure in the image, as can be seen by the singular gray spots (corresponding to a grayscale value of 255)  in the bottom-left magnified cutout of Figure~\ref{fig:preprocessing} (left). It is important to remove these bright spots because, during the training of a CNN, they can strongly influence convolutions and lead to overfitting. This  is implemented by determining all pixels with intensity value equal to 255. For each such pixel $\voxel \in \window$ with $I(\voxel)=255$, we compute the average grayscale value of its neighboring pixels,  where a pixel is considered a neighbor if it is horizontally, vertically, or diagonally adjacent to the pixel under consideration. If the average grayscale value of these neighboring pixels is below a threshold value of 128, we replace $I(\voxel)$ by the calculated average value.

In summary, we applied several preprocessing steps to the WALS image data in order to reduce the input size, while retaining all relevant information and improving the quality of the images for further analysis. From this point on, all analyses and model training steps are performed exclusively on this preprocessed WALS image data.

\subsubsection{Deep neural network architecture}

\begin{figure}[ht!]
    \centering
 \includegraphics[width=1\textwidth]{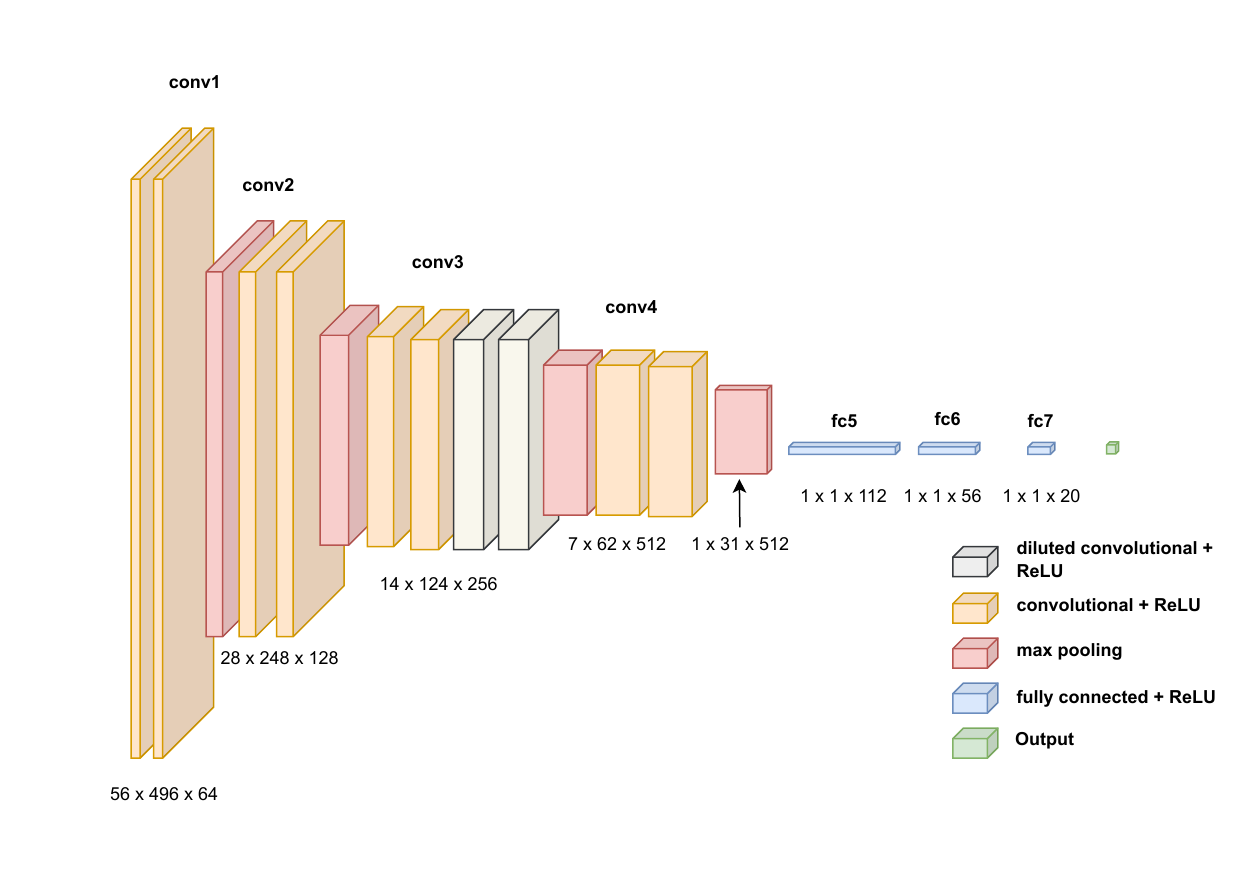}
    \caption{Schematic representation of the deep neural network architecture.}
    \label{fig:network}
\end{figure}

It is important to reiterate that two different networks are used in order to predict the class labels $s_1,s_2$ and the quantitative descriptors of droplet size $s_3,s_4$.  Nonetheless, both networks possess the same basic architecture, which consists of two main parts: a convolutional part and a dense part. The convolutional part is designed to extract and process features from the input data, while the dense part is used to perform classification on the extracted features. A schematic representation of the architecture is shown in Figure~\ref{fig:network}.

The convolutional part of the architecture includes several residual blocks and a dilated residual block, similar to the downsampling part in the U-net architecture used in~\cite{orkun2022l}. The residual blocks are designed to learn and extract features from the input data. Each residual block consists of several convolutional layers with a specified number of filters, followed by batch normalization and rectified linear unit (ReLU) activation layers. The input of each residual block is added to its output in order to form a residual connection, allowing the residual block to learn residual mappings between the input and output, see~\cite{heres}.  The use of such residual blocks has been shown to improve learning in deep neural networks, see~\cite{groceri}. The residual blocks also include a max-pooling layer with a specified pool size.  These max-pooling layers are used to reduce the spatial dimensions of the output from each block, as in~\cite{Unet}. The dilated residual block is designed to increase the receptive field of the network without increasing the number of parameters. The dilated residual block consists of several dilated convolutional layers with a specified dilation rate and number of filters, each of which are followed by batch normalization and ReLU activation layers. 

The dense part of the architecture is used to perform classification and prediction on the extracted features of the convolutional part. The output of the final convolutional layer is flattened and passed through three dense layers with a specified number of units and ReLU activation functions. The final layer, i.e., the output layer, of the base architecture consists of two units, where the activation function is either a sigmoid (for the architecture which determines the class labels $s_1$ and $s_2$) or a ReLU function (for the architecture which predicts the droplet sizes $s_3$ and $s_4$).

\subsubsection{Training procedure}

In order to accurately predict the presence and size of droplets from preprocessed WALS images, we need to train our neural networks on datasets of labeled examples. This training process involves adjusting the network parameters to minimize the discrepancy between the predicted and true outputs.

The training data, denoted by $D$, consists of $n$ pairs of input images and their corresponding output vectors for some integer $n>1$, 
i.e., $D=\{(I_p^{(i)},s^{(i)})\}_{i=1}^n$ where $I_\mathrm{p}^{(i)}$ is the $i$-th input image and $s^{(i)}=(s_1^{(i)},s_2^{(i)},s_3^{(i)},s_4^{(i)})$ is the corresponding ground truth vector. We use two neural networks: one for predicting the binary labels $s_1,s_2$, which indicate whether a droplet was detected on the left and right ring-shaped segments, respectively; and another one for predicting the quantitative descriptors $s_3,s_4$ of droplet size, corresponding to the ground truth droplet size measured in the left and right ring-shaped segments, respectively. Thus, the respective training sets $D_1$ and $D_2$ for the two networks consist of the following  pairs of input images and their corresponding output vectors: $D_1=\{(I_\mathrm{p}^{(i)},(s^{(i)}_1,s^{(i)}_2)): \ i\in\{1,2,\ldots,n \}\}$ and $D_2=\{(I_\mathrm{p}^{(i)},(s^{(i)}_3,s^{(i)}_4)):  i\in\{1,2,\ldots, n\} \text{ with } s^{(i)}_1+s^{(i)}_2>0\}\}$.  The condition $s^{(i)}_1+s^{(i)}_2>0$ for $D_2$ ensures that the training set for the droplet size prediction network includes only those images for which a droplet size can be detected on at least one ring-shaped segment. In order to increase the variety in the training data, we use training data augmentation~\cite{preprocessing,bestpractices}. This involves the application of random transformations to the input images in order to generate additional training data. Specifically, we use reflection and translation as options for data augmentation.  

Reflection involves flipping the image vertically with a probability of 0.5. If the image is flipped vertically, the ground truth values are adjusted accordingly. This adjustment is necessary due to the polar transformation applied to the original image. In this polar transformation, the $y$-axis of the transformed image corresponds to the angular coordinate in the polar system, which is defined with respect to a reference axis drawn from the center to the top edge of the original image. Therefore, a vertical flip of the transformed image is equivalent to a horizontal flip of the original image in Cartesian coordinates. Because the ground truth labels are assigned specifically to the left and right ring-shaped segments of the WALS image, this change in orientation necessitates an adjustment of the ground truth values to maintain the correct association of the labels. By reflecting a WALS image, we essentially create a mirror image of the droplet, i.e., as if it were captured in a different spot. Therefore, applying reflections during training improves the model's ability in detecting droplets consistently, no matter where they appear in the measurement area.
For translation, we shift the image in both the $x$- and $y$-directions by a maximum of two pixels which corresponds to an angular change of approximately 1.5 degrees in the original WALS image. Importantly, all movements within this range are equally likely, ensuring a uniform distribution of translations for the data augmentation process. This method is chosen to train models that remain consistent despite minor variations within the experimental setup. This characteristic is crucial for the dependable and practical analysis of future experimental data.

During the training phase for each neural network, we employ a mini-batch gradient descent approach in order to update the networks parameters. More specifically, for a given batch size \( b \le n \), we repeatedly select  batches of \( b \) random images from their designated training sets and input them into the corresponding neural network, producing predictions. These predictions are then compared to the ground truth labels and the discrepancy is quantified via a specific loss function. Subsequently, the gradient of this loss function with respect to the network parameters is computed to refine the model. Based on this gradient, adjustments to the parameters are made using the Adam optimizer, as referenced in~\cite{adam}. This optimizer adapts the learning rate for each parameter depending on the first and second moments of the gradients.

This iterative process of batch selection, prediction generation, loss computation, and parameter updating is conducted over several epochs. In an initial experiment using image data at a HAB of 100 mm, convergence of training and validation loss was observed around the 50-epoch mark. There was no significant rise in validation loss in subsequent epochs, suggesting a lack of overfitting. Given the considerable data volume across all HABs, a total of 100 training epochs, each with 100 batches, was deemed appropriate, as supported by the trends shown in Figure~\ref{fig:loss_vs_epochs}, demonstrating a balance between accuracy and computational cost. 

During network training, the quality of our predictions is continuously assessed. This assessment is primarily driven by the loss functions, which highlight the difference between the predictions and corresponding ground truth values. In the next section, the loss functions considered in this paper are described in detail.

\subsubsection{Loss function}

To train our neural networks in order to accurately predict the binary labels $s_1,s_2$ and the quantitative descriptors of droplet size $s_3,s_4$, we need to consider suitably chosen loss functions. They measure the discrepancy between the predicted values $\widehat{s}_j$ and the true values $s_j$, for $j=1,2,3,4$, providing a feedback mechanism for the neural networks to adjust their parameters during training. Given the distinct nature of our two tasks - one of them performing classification and the other one performing qunatitative  scalar prediction - we employ different loss functions for each case. This allows us to tailor our approach according to the specific requirements of each task.

For training the droplet detection network, we use the mean absolute error (MAE) as loss function, where the MAE calculates the mean absolute difference between the first two components of the predicted labels $\widehat s=(\widehat{s}_1,\ldots,\widehat{s}_4)$  and the corresponding ground truth labels $s=({s}_1,\ldots,{s}_4)$. Specifically, for a set of $b$ predictions $\widehat{s}^{(1)},\widehat{s}^{(2)},\ldots,\widehat{s}^{(b)}$ and corresponding ground  truth labels $s^{(1)},s^{(2)},\ldots,s^{(b)}$, the MAE for the $j$-th component is given by
\[
\text{MAE}_j = \frac{1}{b} \sum_{i=1}^{b} |\widehat{s}_j^{(i)} - s_j^{(i)}| \qquad\text{ for } j\in\{1,2,3,4\}.
\]

Moreover, for the training of the droplet detection network, we consider both \( \text{MAE}_1 \) and \( \text{MAE}_2 \) as they relate to the binary labels for droplet detection on the left and right ring-shaped segments, respectively. Therefore, the training loss for the detection network is the average of \( \text{MAE}_1 \) and \( \text{MAE}_2 \), i.e., the value of $({\text{MAE}_1+\text{MAE}_2})/2$.

For the droplet size estimation network, using the MAE is not optimal because the relative impact of an absolute error is much more pronounced for smaller particles. Therefore, we need a loss function that equally emphasizes the importance of errors across all particle sizes during training. This can be achieved by a loss function which is based on the symmetric mean absolute percentage error (SMAPE). The SMAPE calculates the absolute percentage difference between predicted labels  \(\widehat{s}\) and the ground truth \(s\). Specifically,  the SMAPE  for the $j$-th component is defined as
\[
\text{SMAPE}_j = \frac{1}{b} \sum_{i=1}^{b} \frac{2|\widehat{s}_j^{(i)} - s_j^{(i)}|}{|\widehat{s}_j^{(i)}| + |s_j^{(i)}|}\qquad \text{ for } j\in\{3,4\}.
\]

As already mentioned above, the advantage of using the SMAPE for the droplet size estimation network is that it is a scale-invariant measure of accuracy. This means that it assigns equal importance to relative errors, regardless of the magnitude of the true values. This property is particularly useful when dealing with data that spans a wide range, such as droplet sizes.

However, as outlined in Section~\ref{sec:gt}, our approach for the generation of ground truth labels involves analyzing both ring-shaped segments of the WALS image separately. Consequently, a significant number of images in our dataset only has ground truth values for droplet size on one ring-shaped segment, e.g., $s_1^{(i)}=1$ and $s_2^{(i)}=0$, in which case we can calculate a meaningful size discrepancy for the left ring-shaped segment, but not for the right one. To effectively utilize this data as well, we employ a weighted SMAPE loss function, where the weights are used to exclude data points without ground truth droplet size values by multiplying the computed discrepancy with a weight of $0$.  This ensures that the SMAPE is computed only for the ring-shaped segments of the WALS image in which droplets were detected. As a result, this approach enables us to accurately evaluate the performance of our size estimation network on all available data.  The weights of $1$ and $0$ correspond to the cases  whether or not a droplet was detected in the ground truth. For example, if $s_1^{(i)}=1$ then  $s_3^{(i)}$ is a valid droplet size. 
Therefore, the weighted SMAPE ($\text{SMAPE}_{\text{w}}$) of the $j$-th component is given by  

$$ \text{SMAPE}_{\text{w},j} =  \frac{1}{b}\sum_{i=1}^{b} s_{j-2}^{(i)}\frac{2 |\widehat{s}^{(i)}_j - s^{(i)}_j|}{|\widehat{s}^{(i)}_j| + |s^{(i)}_j|} \qquad\text{ for } j\in\{3,4\}. $$ 

Again, finally, we consider the average of the losses computed on the left and right ring-shaped segments, i.e., the value of  \( (\text{SMAPE}_{\text{w},3}+\text{SMAPE}_{\text{w},4})/2 \).

This combination of the loss functions described above allows us to accurately measure the performance of the neural networks while taking into account all available data.

\section{Results}\label{sec.the}

In this section, we present the results of our analysis for the characterization of droplets using WALS images. We evaluate the performance of two network models: one for droplet detection and one for droplet size estimation. In order to ensure that our models are able to generalize to new data, we consider a certain cross-validation process. For that purpose, different pairs of droplet detection and droplet size prediction models were trained for each HAB, using only the data from the other HABs. We then evaluate these models on the data from the HAB that was not used for training. This cross-validation process is repeated for all HABs.
More precisely, let 
$
H = \{20,  30, \ldots, 120\}
$

denote the set of all HABs. Then, for each HAB $h \in H$, we train two models with data associated with the remaining HABs  from $H\setminus \{h\}$. The models trained in this way, we denote by $M_{1,h}$ and $M_{2,h}$, respectively, where $M_{1,h}$ denotes the  model for droplet detection and $M_{2,h}$ is the model for the prediction of droplet sizes. Note that the results presented in this section are  averaged across all HABs, unless otherwise specified. Thus, if for $i=1,2$ we consider some function $\Phi_i(h)$ which assigns the model $M_{i,h}$ some performance measure that has been computed on the HAB $h\in H$, 
 then we generally consider the cross-validation value $\Phi_i$ given by
$$
\Phi_i= \frac{1}{\#H} \sum_{h \in H} \Phi_i(h),
$$
where $\#H$ denotes cardinality of the set $H$, i.e., $\#H=11$ in our case.

\subsection{Model performance for different numbers of  training epochs}

\begin{figure}[ht] \centering \begin{subfigure}{0.45\textwidth} \centering \caption{} \includegraphics[width=\textwidth]{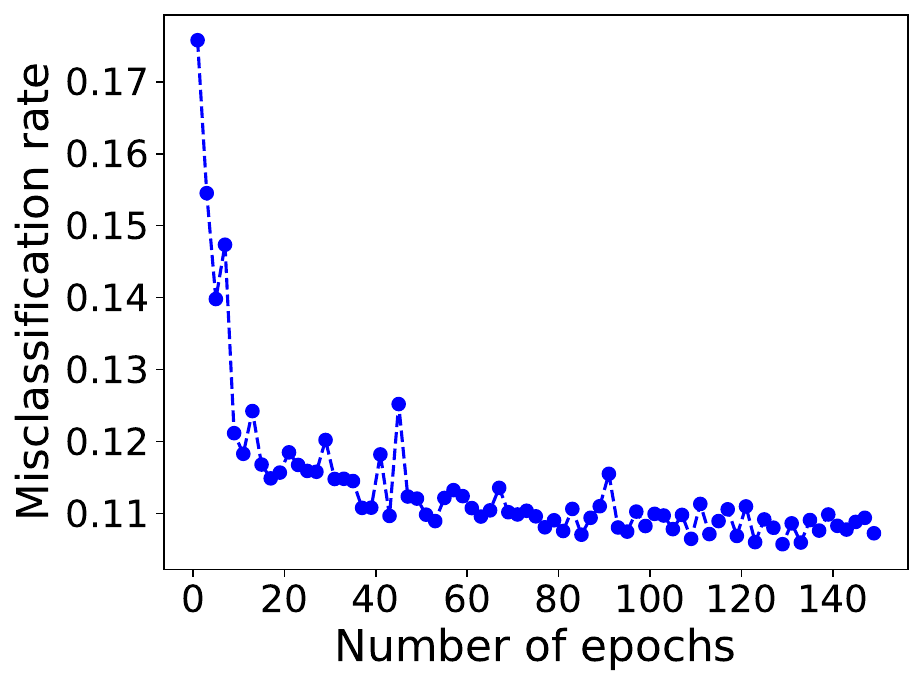} \label{fig:loss_vs_epochs_1} \end{subfigure}\hfill \begin{subfigure}{0.45\textwidth} \centering \caption{} \includegraphics[width=\textwidth]{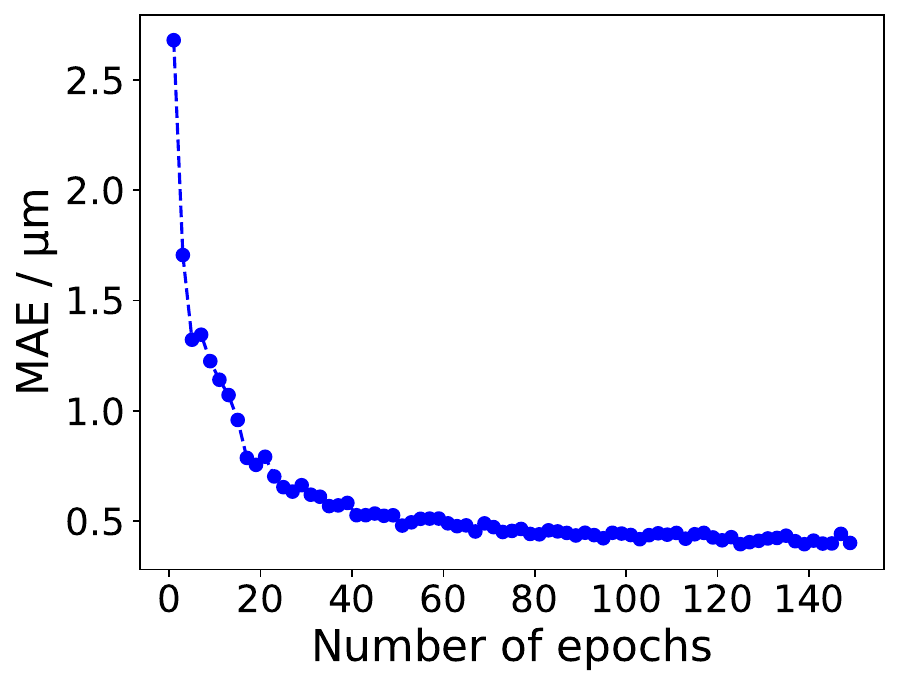} \label{fig:loss_vs_epochs_2} \end{subfigure} \vspace{-10pt} \caption{Values of performance measures for different numbers of training epochs for the droplet detection model (left) and the droplet size estimation model (right).} \label{fig:loss_vs_epochs} \end{figure}

Figure~\ref{fig:loss_vs_epochs} shows  plots of  model performance for different numbers of  training epochs for the two network models considered in this paper, where the  performance of the droplet detection model is quantified by the classification rate, i.e., the probability that a WALS image is misclassified. The performance of the droplet size estimation model is quantified by the MAE of the predicted droplet sizes.

 Note that, while the SMAPE is used during training to ensure that errors are considered in relation to droplet size, the MAE is a more interpretable performance measure which is commonly used for visualization purposes. As it can be seen in Figure~\ref{fig:loss_vs_epochs}, the performance improves  rapidly until training epoch 20, after which the rate of improvement slows down. However, both models continue to improve their performance until epoch 100, indicating that choosing a training length of 100 epochs is reasonable. We then obtained that the model for droplet detection exhibits a misclassification rate of approximately 11\%, while the model for droplet size estimation has an MAE of approximately 0.5$\,\mu$m.

\begin{figure}[ht]
\centering
\includegraphics[width=0.5\textwidth]{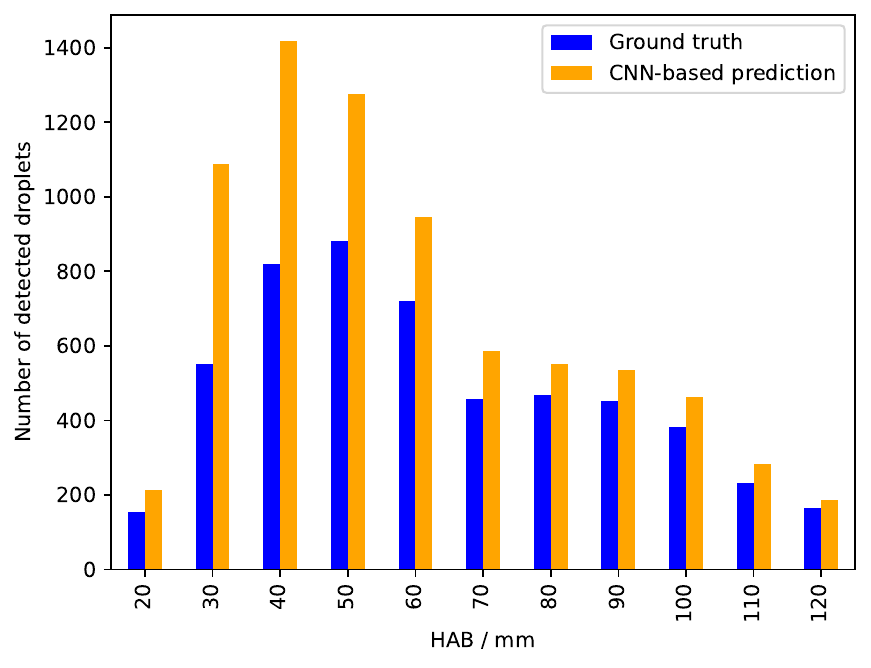}
\caption{Measured and predicted numbers of droplets for each HAB from the set $H=\{20,30,\ldots,120\}$ at a training epoch of 100.}
\label{fig:histogram_125}
\end{figure}

\subsection{Misclassification rate and droplet size  across different HABs}

Furthermore, we performed an analysis of the classification rate across different HABs.
Figure~\ref{fig:histogram_125} shows large discrepancies between the measured and predicted numbers of  droplets at three HABs (30, 40, 50), indicating that the model predicts a substantially larger number of droplets at certain heights.

However, it is possible that these additional droplets detected by the model, but absent in the ground truth data, follow a size distribution which is consistent with that of the droplets in the ground truth.
 Therefore, we  analyzed  the impact of these additional droplets on the resulting size distribution.

For that purpose, it is important to recall that the images have individual labels for both the left and right ring-shaped segments, assigning a droplet size to the corresponding ring-shaped segment if a droplet was detected. In order to consolidate the droplet sizes from both ring-shaped segments into a single value per image corresponding to the size of the measured droplet, we consider the  quantity $S_{\text{D}}$, which takes the four-dimensional ground truth vector $s=(s_1,s_2,s_3,,s_4)$ as input, where
\begin{equation}
\label{eq:size}
S_\text{D} = \begin{cases}
\text{-1,} & \text{if } s_1 = 0 \text{ or } s_2= 0, \\
\text{-1,} & \text{if } s_1 > 0,  s_2> 0 \text{ and }\displaystyle \frac{|s_3 - s_4|}{\max\{s_3,s_4\}} > 0.15, \\
\displaystyle
\frac{s_3+s_4}{2}, & \text{otherwise.}
\end{cases}
\end{equation}

In addition, this formula is  applied to the corresponding predicted four-dimensional   vector $\widehat s=(\widehat s_1,\widehat s_2,\widehat s_3,\widehat s_4)$ to determine the predicted droplet size $\widehat{S}_\text{D}$. Formula~\eqref{eq:size} specifies that if a droplet is not detected on both ring-shaped segments  or if the estimated sizes from each ring-shaped segment disagree by more than 15\%, then no droplet is considered to have been measured, as indicated by the fictitious value $-1$. The threshold of 15\% has been selected heuristically and can be adjusted for new data sets if necessary.

\begin{figure}[ht]
    \centering
    \includegraphics[width=\textwidth]{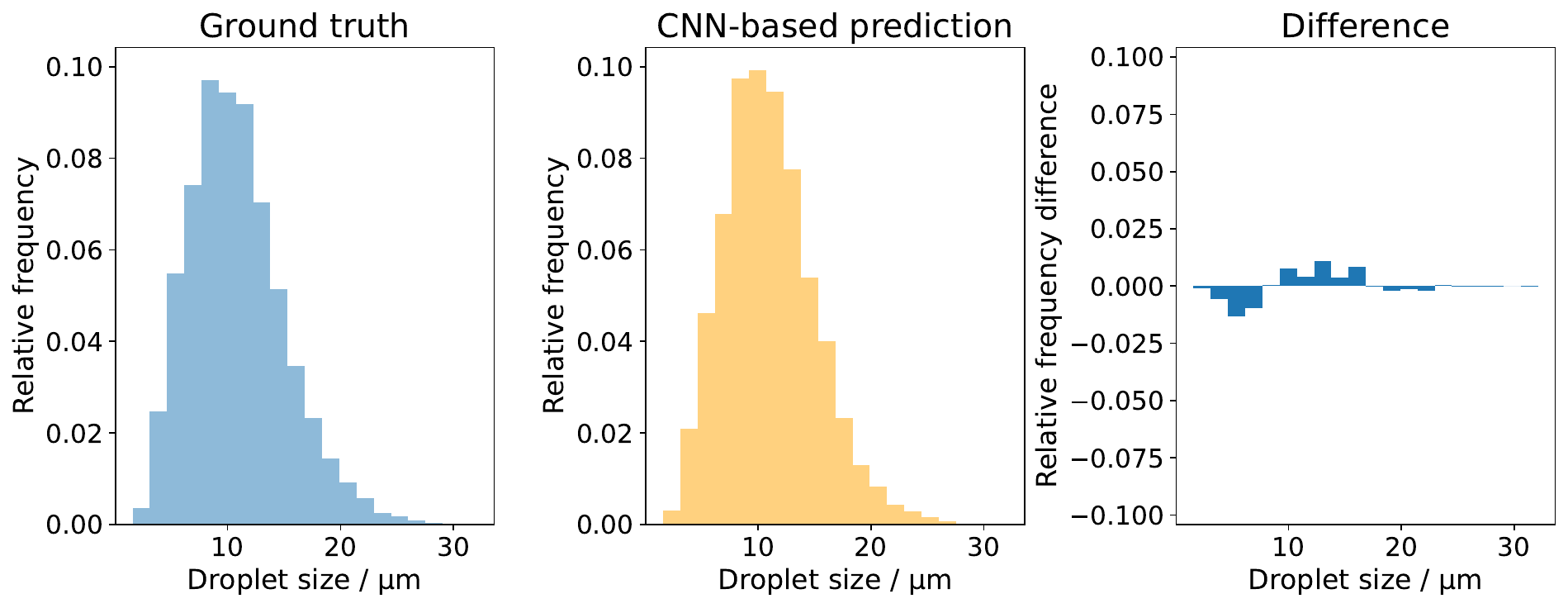}
    \caption{Droplet size histograms for ground truth (left), predictions (middle) and difference (right) across different HABs.}
    \label{fig:droplet_size_distributions}
\end{figure}

As shown in Figure~\ref{fig:droplet_size_distributions}, the histograms of droplet size determined by means of Equation~\eqref{eq:size} for ground truth (left) and predictions (middle)  are quite similar. To further quantify this similarity, we compared the median and interquartile range (IQR) of both the ground truth and  predicted droplet sizes. Note that the median provides a measure of the central tendency of the distribution, while the IQR provides a measure of its spread, where we obtained a median  of 10.4$\,\mu$m  and an IQR  of 5.5$\,\mu$m for the ground truth data, and  a median  of 10.6$\,\mu$m and an IQR value of 5.4$\,\mu$m for the predicted distribution. This indicates that both the central tendency and spread of the two distributions are similar, suggesting that the additional droplets detected by the prediction model do not significantly alter the resulting size distribution.

\subsection{Comparison of two different methods for the prediction of droplet sizes}

Moreover, we compared the CNN-based approach considered in the present paper with another automatic  method for droplet size prediction, see Figure~\ref{fig:boxplot}. 
The latter method, which we refer to as the "baseline method," considers the droplet characterization process as described in Section~\ref{sec:gt}, however, without the manual removal of erroneously detected WALS images,  see~\cite{Assmann2021}.  In particular, this means that the baseline method  predicts the same  droplet sizes for those WALS images which did not need to be manually removed. Therefore, when the necessity for manual removal of images is low, the baseline method can provide a reasonable estimation of droplet size.

On the other hand, the baseline method lacks the ability to discern and exclude erroneously detected WALS images, which can lead to inaccurate predictions of droplet size. Therefore, when the data contains many such erroneously detected  images, the performance of the baseline method can be significantly impaired and thus impact subsequent analyses.

\input{boxplot_latex}

Comparing the performance of the CNN-based approach with that of the baseline method is essential as it allows us to assess the improvements brought by the incorporation of CNNs in droplet size prediction. Furthermore, by comparing  results of the present paper with those obtained by  the baseline method at each individual HAB, we  better understand how the performance of the CNN-based approach varies under the different measurement conditions associated with each HAB.
The results of this analysis are depicted in Figure~\ref{fig:boxplot}, which provides a visual comparison of the droplet sizes determined by the ground truth, the CNN-based predictions, and the baseline method at different HABs. Each boxplot represents the median, IQR, and outliers for each respective set of data.

\subsection{Parametric distributions of droplet sizes}

To further analyze the similarity between the ground truth and CNN-based prediction of droplet sizes, we fit parametric distributions to both datasets. 
Specifically, we consider the log-normal distribution, which is commonly used to model droplet
size distributions. Note that the probability density function of the log-normal distribution is given by
$$
f(x) = \frac{1}{x\ln\sigma_{\mathrm{g}}\sqrt{2\pi}}{\rm e}^{-\frac{(\ln x - \ln \mu_{\mathrm{g}})^2}{2(\ln \sigma_{\mathrm{g}})^2}},\qquad\text{for each }  x>0,
$$
where $\mu_{\mathrm{g}}, \sigma_{\mathrm{g}}>0$ are model parameters. The model parameters $\mu_{\mathrm{g}} \text{ and } \sigma_{\mathrm{g}}$ correspond to the geometric mean and the geometric standard deviation of the resulting probability distribution, respectively. We determined the values of these parameters using maximum likelihood estimation and compared the values obtained for the ground truth data and the CNN-based predictions, respectively, see Table~\ref{tab:lognorm}.

\input{Images/table2}

 Moreover,  in  Figure~\ref{fig:lognormfits}, log-normal probability densities of droplet sizes are shown, where it is clearly visible that, for the measuring height of 30 mm,  the probability density fitted to CNN-based predictions approximates the probability density of ground truth data better than the probability density fitted to data obtained by the baseline method. When data from all available HABs is considered, both the CNN-based approach as well as the baseline method lead to accurate approximations of the log-normal probability density computed from ground truth data, see Figure~\ref{fig:lognormfits}.

\input{figure4}

\section{Discussion}\label{sec.fou}

 The obtained results indicate that the model for droplet detection exhibits a misclassification rate of approximately 11\%, while the model for droplet size estimation demonstrates high accuracy with an MAE of approximately 0.5$\,\mu$m. These results suggest that the CNN-based approach considered in the present paper is able to accurately detect  individual droplets, and estimate their size, in WALS images.

Our analysis also revealed that the performance of the models varies across different HABs. In particular, three HABs (30, 40 and 50 mm) exhibit large discrepancies between the number of measured and predicted droplets. However, when investigating the resulting droplet size distributions by combining the output of both models using Equation~(\ref{eq:size}), it is indicated that these additional droplets do not significantly alter the resulting size distributions, as shown in Figure~\ref{fig:droplet_size_distributions}. This suggests that the  methodology  presented in this paper is able to accurately capture the underlying size distribution of droplets in WALS images.

Moreover, in Figure~\ref{fig:boxplot} we also evaluated another automatic method for droplet size prediction, which we called the baseline method and which operates similarly to the ground truth method except for the fact that it does not involve manual removal of erroneously detected WALS images. Consequently, the baseline method naturally aligns very closely with the ground truth at HABs where almost no manual removal was necessary. 

Despite that, at HABs of 30, 40, and 50 mm, the droplet sizes predicted by the CNN-based approach resulted in medians and IQRs that were closer to the ground truth than those resulting from the droplet sizes predicted by the baseline method. These differences are visualized by the vertical line (median) and box boundaries (IQR) in Figure~\ref{fig:boxplot}. Furthermore, Figure~\ref{fig:lognormfits} (left) visualizes this divergence of the results of the baseline method from the ground truth with respect to the resulting fitted log-normal probability densities at the HAB of 30 mm.  
Most importantly, this deviation from the ground truth (of the results obtained by the baseline method at these particular HABs) coincides with a higher rate of manual filtering, as noted in~\cite{Assmann2021}, and provides a context for understanding the performance of the CNN-based approach. 
In particular,  the CNN-based approach considered in the present paper achieves a closer alignment with the ground truth at the HABs of 30, 40, and 50 mm, which indicates its effectiveness in dealing with erroneously detected images. This suggests that the CNN-based approach offers significant improvements over the  baseline method in scenarios requiring extensive manual filtering, underlining the value of incorporating CNNs in droplet size determination.

However, there are several challenges that need to be addressed in future research. One such challenge is the investigation of the robustness of our network models to variations in the amount of training data. Specifically, this could be investigated by withholding large parts of the training data during training. Additionally, we only investigated data coming from a specific experimental setup and it would be desirable to extend our approach to other experimental setups in order to better evaluate the generalization capabilities of the CNN-based approach.

Despite these challenges, our results demonstrate the potential of using machine learning models for droplet characterization from WALS images.

\section{Conclusion and outlook}\label{sec.fiv}
When investigating the formation of hetero-aggregates, the accurate characterization of large amounts of droplets is important in order to refine the synthesis processes of hetero-aggregates. Therefore, in this study, we presented a fully automatic method for droplet characterization from WALS images using machine learning models.

The obtained results indicate that the  CNN-based approach proposed in this paper is able to accurately detect droplets and estimate their sizes from  WALS images. The accuracy of this CNN-based approach was consistent across all investigated HABs in the range of 30 mm to 120 mm, which was not the case for a state-of-the-art approach for automatic droplet size characterization from WALS images, which we used as a benchmark. However, we also identified several challenges that need to be addressed in future research.

One promising direction for future research is the generation of synthetic data with known ground truth to train our models. This could potentially bypass the need for manual labeling of training data and allow for a more systematic investigation of model performance. Moreover, when the density of measured droplets is high (as it is the case for smaller HABs), multiple droplets are present in the WALS probe volume at the same time, leading to overlapping scattering data. The resulting images are currently classified as erroneous images in the ground truth, whereas a combination of synthetic data and machine learning could potentially estimate accurate droplet sizes in those cases as well.

Overall, our results demonstrate the potential of using machine learning models for droplet characterization from WALS images. These results are significant for enhancing droplet analysis techniques through the application of machine learning advancements.

\section{Acknowledgements}

Funding by the German Research Foundation (DFG) under grants  \text{WI 1602/16-2}, \text{WI 1602/18-1} and \text{SCHM 997/42-1}.

\medskip

\bibliographystyle{abbrv}

\bibliography{refs}

\end{document}

%% file: false_color.tex
\begin{figure}[h]

\centering

\tikzset{every picture/.style={line width=0.75pt}} 

\begin{tikzpicture}[x=0.75pt,y=0.75pt,yscale=-1,xscale=1]

\draw (309.5,159) node  {\includegraphics[width=185.25pt,height=151.5pt]{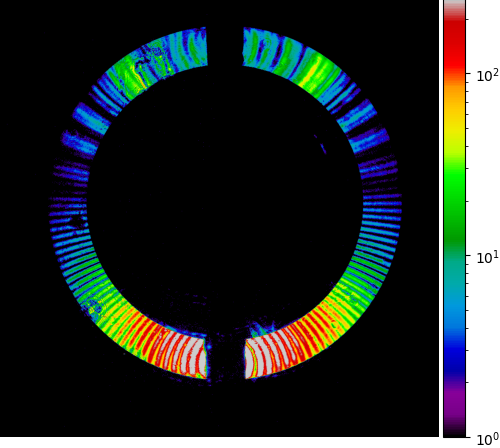}};
\draw (133.1,197.3) node  {\includegraphics[width=77.85pt,height=70.05pt]{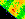}};
\draw [color={rgb, 255:red, 255; green, 255; blue, 255 }  ,draw opacity=1 ][fill={rgb, 255:red, 227; green, 39; blue, 39 }  ,fill opacity=1 ]   (185,244) -- (244,214) ;
\draw [color={rgb, 255:red, 255; green, 255; blue, 255 }  ,draw opacity=1 ][fill={rgb, 255:red, 227; green, 39; blue, 39 }  ,fill opacity=1 ]   (185,150.6) -- (244,208) ;
\draw (506.93,156.64) node  {\includegraphics[width=22.39pt,height=147.54pt]{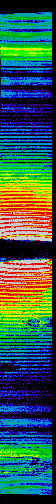}};

\draw (250,36) node [anchor=north west][inner sep=0.75pt]   [align=left] {1000 $\displaystyle \times $ 1000};
\draw (478,36) node [anchor=north west][inner sep=0.75pt]   [align=left] {56 $\displaystyle \times $ 496};
\draw (107,128) node [anchor=north west][inner sep=0.75pt]   [align=left] {25 $\displaystyle \times $ 19};

\end{tikzpicture}

\caption{A side-by-side comparison of the original  WALS data (left) and the preprocessed image (right). The preprocessing involved applying a polar transformation  and removing imaging artifacts (gray spots in the magnified cutout).  The original image has a resolution of $1000\times1000$, while the preprocessed image (right) has a resolution of $56\times496$.}\label{fig:preprocessing}
\end{figure}

%% file: boxplot_latex.tex
\begin{figure}[htbp]
\centering
\includegraphics[width=0.7\textwidth]{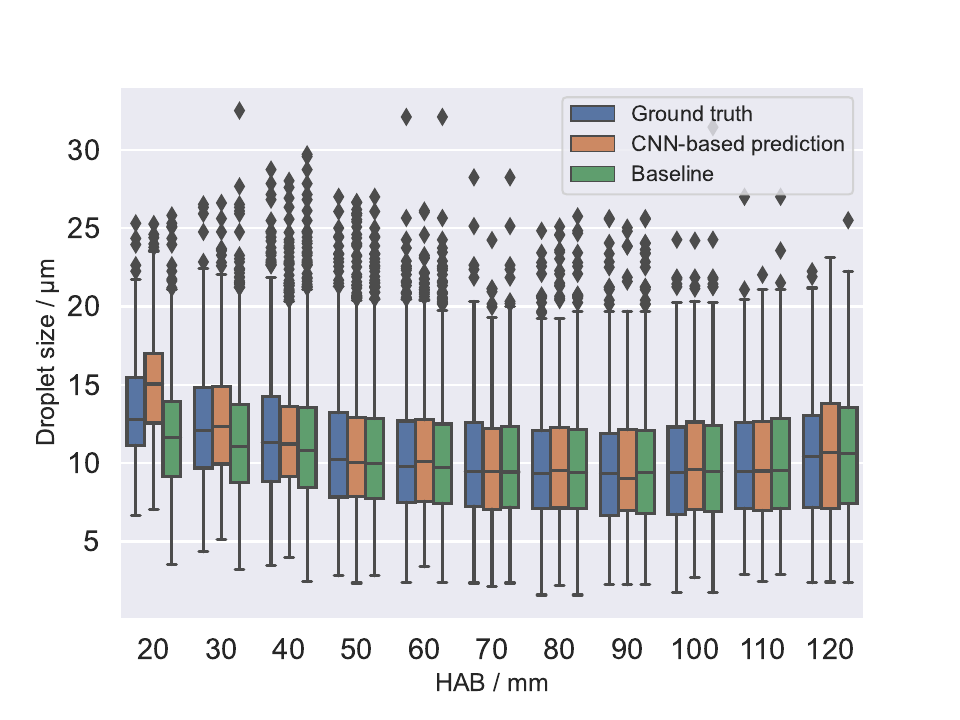}
\caption{Boxplot of droplet sizes at different measuring heights, computed for ground-truth data and two different prediction methods. The boxes represent the IQR of the data. The whiskers extend to the most extreme data points within 1.5 times the IQR from the first and third quartiles. The horizontal line inside each box indicates the median of the data, and any outliers are plotted as individual points beyond the whiskers.
}
\label{fig:boxplot}
\end{figure}

%% file: Images/table2.tex
\begin{table}[ht]
\centering
\caption{Comparison of model parameters. The pairs ($\mu_{\mathrm{gt}},\sigma_{\mathrm{gt}}$) and $(\mu_{\mathrm{pred}},\sigma_{\mathrm{pred}})$ are the  parameter values  fitted to the ground truth and CNN-based prediction of droplet sizes, respectively.}
\label{tab:lognorm}
\begin{tabular}{l|lr|lr}
\toprule
measuring height / mm  &  $\mu_{\mathrm{g,gt}}$ / $\mu$m  &  $\sigma_{\mathrm{g,gt}}$ &  $\mu_{\mathrm{g,pred}}$ / $\mu$m  &  $\sigma_{\mathrm{g,pred}}$ \\
  \hline         
\midrule
20               &        12.9 &           1.31 &          14.7 &             1.29 \\
30               &        11.9 &           1.36 &          12.1 &             1.33 \\
40               &        11.1 &           1.44 &          11.1 &             1.36 \\
50               &        10.1 &           1.47 &          10.0 &             1.46 \\
60               &         9.7 &           1.49 &           9.8 &             1.47 \\
70               &         9.2 &           1.52 &           9.0 &             1.52 \\
80               &         9.1 &           1.53 &           9.2 &             1.54 \\
90               &         8.9 &           1.57 &           9.0 &             1.54 \\
100              &         9.0 &           1.56 &           9.2 &             1.53 \\
110              &         9.1 &           1.53 &           9.1 &             1.56 \\
120              &         9.2 &           1.62 &           9.5 &             1.64 \\
Aggregated       &        10.0 &           1.51 &          10.2 &             1.48 \\

\end{tabular}

\end{table}

%% file: figure4.tex
\begin{figure}[ht]
\centering
\begin{subfigure}{0.45\textwidth}
\includegraphics[width=\textwidth]{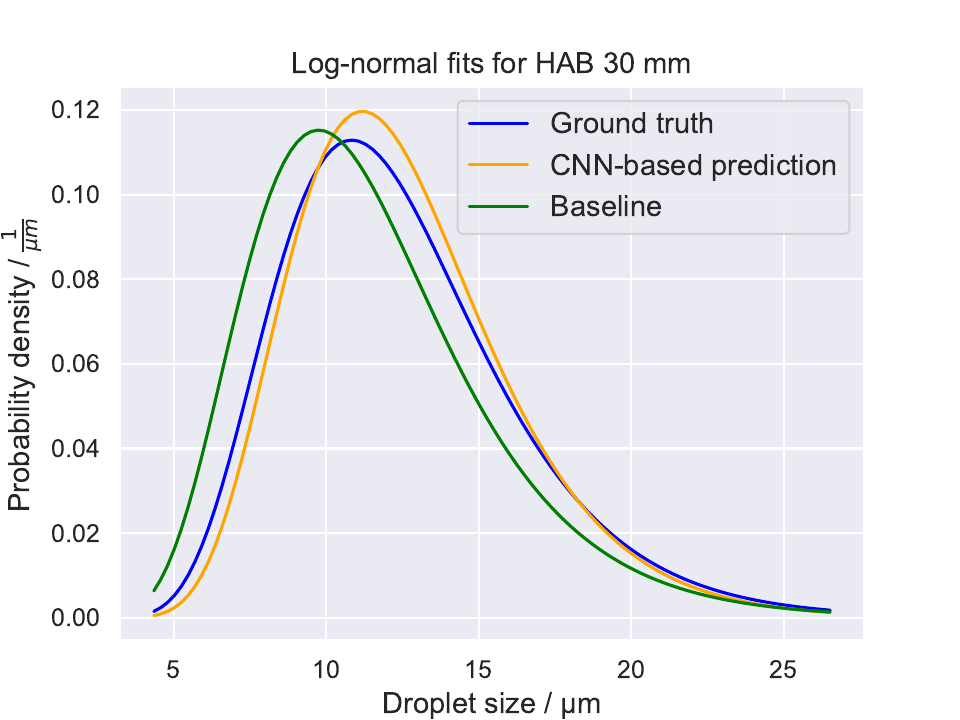}
\end{subfigure}
\hfill
\begin{subfigure}{0.45\textwidth}
\includegraphics[width=\textwidth]{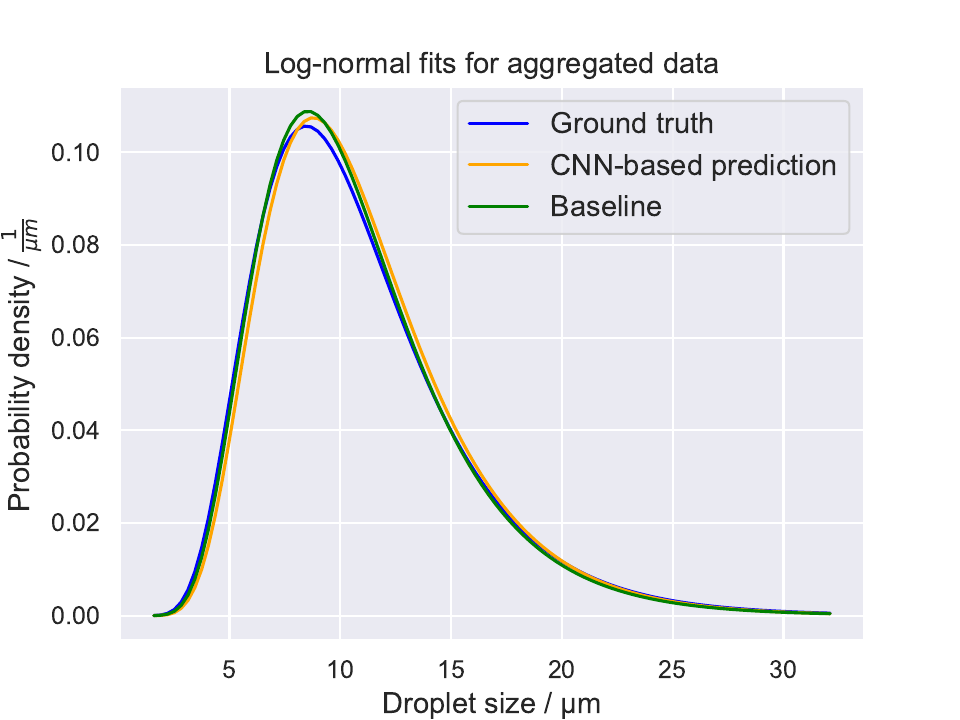}
\end{subfigure}
\caption{Comparison of log-normal droplet size densities,  for the measuring height of 30 mm (left), and for the aggregated data across all measuring heights (right). 
}\label{fig:lognormfits}

\end{figure}